# Benchmarks, Performance Evaluation and Contests for 3D Shape Retrieval

Afzal Godil[1], Zhouhui Lian[1], Helin Dutagaci[1], Rui Fang[2], Vanamali T.P. [1], Chun Pan Cheung[1]
[1]National Institute of Standards and Technology, USA
[2]School of Computer Science and Engineering, Michigan State University, USA
godil@nist.gov

## ABSTRACT
Benchmarking of 3D Shape retrieval allows developers and researchers to compare the strengths of different algorithms on a standard dataset. Here we describe the procedures involved in developing a benchmark and issues involved. We then discuss some of the current 3D shape retrieval benchmarks efforts of our group and others. We also review the different performance evaluation measures that are developed and used by researchers in the community. After that we give an overview of the 3D shape retrieval contest (SHREC) tracks run under the EuroGraphics Workshop on 3D Object Retrieval and give details of tracks that we organized for SHREC 2010. Finally we demonstrate some of the results based on the different SHREC contest tracks and the NIST shape benchmark.

## 1. INTRODUCTION
3D objects are widespread and present in many diverse fields such as computer graphics, computer vision, computer aided design, cultural heritage, medical imaging, structural biology, and other fields. Large numbers of 3D models are created every day using 3D modeling programs and 3D scanners and many are stored in publicly available databases. These 3D databases require methods for storage, indexing, searching, clustering, and retrieval to be used effectively. Hence, content based 3D shape retrieval has become an active area of research in the 3D community. Benchmarking allows researchers to evaluate the quality of results of different 3D shape retrieval approaches. Under a benchmark, different shape matching algorithms are compared and evaluated in term of efficiency, accuracy, robustness and consistence. Results are then obtained and conclusions of the performance are drawn towards the shape matching algorithms.

In section 2, the related work of previous benchmarks is briefly reviewed; in section 3, we discuss benchmarks and the construction of the benchmark; in section 4, we present the evaluation measures used; the NIST shape benchmark is discussed and analyzed in section 5; section 6 describes the SHREC contests and their results; and finally conclusions are drawn in section 7.

## 2. RELATED WORK
***Contest*** The SHape REtrieval Contest (SHREC) [4] is organized every year since 2006 by Network of Excellence AIM@SHAPE under the EuroGraphics Workshop on 3D Object Retrieval to evaluate the effectiveness of 3D shape retrieval algorithms. In 2006, one track was organized to retrieve 3D mesh models on the Princeton Shape Benchmark [1]. In the SHREC 2007, several tracks were organized which focused on specialized problems: the watertight models track, the partial matching track, the CAD models track, the protein models track, the 3D face models track. In the SHREC 2008, following tracks are organized: stability of watertight models, the track on the classification of watertight models and the generic models track, 3D face models. In the SHREC 2009, there were four tracks organized, and we organized two tracks, one based on Generic shape retrieval and the other based on partial shape matching. For the SHREC 2010, there were 11 tracks organized initially, two of them were cancelled because of not enough participants and we organized three Shape Retrieval tracks: Generic 3D Warehouse; Non-Rigid Shapes; and Range Scans.

***Benchmark*** One of the main 3D Shape Retrieval benchmarks is the Princeton Shape Benchmark [1], which is a publicly available database of 3D polygonal models with a set of software tools that are widely used by researchers to report their shape matching results and compare them to the results of other algorithms. The Purdue engineering shape benchmark [2] is a public 3D shape database for evaluating shape retrieval algorithms mainly in the mechanical engineering domain. The McGill 3D shape benchmark [3] provides a 3D shape repository which includes models with articulating parts. Other current shape benchmarks were introduced and analyzed in [10]. We also have developed two Generic shape benchmarks [6], [7] and a Range scan benchmark [8] which we hope will provide valuable contributions to the 3D shape retrieval and evaluation community.

## 3. BENCHMARKS
In this section, the benchmark design principles and how to build the ground truth for benchmarks are discussed, respectively.

### 3.1 Benchmark Design Principles
A number of issues need to be addressed in order to create a 3D Shape benchmark dataset. The dataset must be available free of charge and without copyright issues, so the dataset can be located on a website and can be used freely by everyone for publications. The issue is getting a large collection of 3D models that maybe freely used, which includes those in the public domain, and also ones that are freely licensed, like under the GNU Free Doc. license, or some of the Creative Commons licenses, which offers the Authors/Artists alternatives to the full copyright. There are two main steps to benchmark a shape database, the first of which is to get enough 3D shape models. All the 3D models in the new shape benchmark were acquired by the web crawler. The other step is to classify the 3D shape models into a ground truth database; we discuss it below in detail. 3D models down-loaded from websites are in arbitrary position, scale and orientation, and some of them have many types of mesh errors. Shapes should be invariant to rotation, translation and scaling, which require the process of pose normalization before many shape descriptors can be applied to extract shape features.

## 3.2 Building a Ground Truth for Benchmark

The purpose of benchmarking is to establish a known and validated ground truth to compare different shape matching algorithms and evaluate new methods by standard tools in a standard way. Building a ground truth database is an important step of establishing a benchmark. A good ground truth database should meet several criteria [12], like, having a reasonable number of models, being stable in order to evaluate different methods with relatively high confidence, and having certain generalization ability to evaluate new methods. To get a ground truth dataset, in text retrieval research, TREC [5] uses pooling assessment [12]. In image retrieval research, as there is no automatic way to determine the relevance of an image in the database for a given query image [18], the IAPR benchmark [11] was established by manually classifying images into categories. In image processing research, the Berkeley segmentation dataset and benchmark [14] assumes that the human segmented images provide valid ground truth boundaries, and all images are segmented and evaluated by a group of people. As there is no standard measure of difference or similarity between two shapes, in our shape benchmark [6], two researchers were assigned as assessor to manually classify objects into ground truth categories. When there are disagreements on which category some objects should belong, another researcher was assigned as the third assessors to make the final decision. This classification work is purely according to shape similarity, that is, geometric similarity and topology similarity. Each model was input to a 3D viewer, and the assessor rendered it in several viewpoints to make a final judgment towards shape similarity.

# 4. EVALUATION MEASURES

The procedure of 3D shape retrieval evaluation is straightforward. In response to a given set of users' queries, an algorithm searches the benchmark database and returns an ordered list of responses called the ranked list(s), the evaluation of the algorithm then is transformed to the evaluation of the quality of the ranked list(s). Next, we will discuss the evaluation method that we have used.

Different evaluation metrics measure different aspects of shape retrieval behavior. In order to make a thorough evaluation of a 3D shape retrieval algorithm with high confidence, we employ a number of common evaluation measures used in the information retrieval community [12].

## 4.1 Precision- Recall

Precision- Recall Graph [12] is the most common metric to evaluate information retrieval system. Precision is the ratio of retrieved objects that are relevant to all retrieved objects in the ranked list. Recall is the ratio of relevant objects retrieved in the ranked list to all relevant objects.

Let A be the set of all relevant objects, and B be the set of all retrieved objects then,

$$precision = \frac{A \cap B}{B} \quad \text{and} \quad recall = \frac{A \cap B}{A} \qquad (1)$$

Basically, Recall evaluates how well a retrieval algorithm finds what we want and precision evaluates how well it weeds out what we don't want. There is a tradeoff between Recall and Precision, one can increase Recall by retrieving more, but this can decrease Precision.

## 4.2 R-precision

The precision score when R relevant objects are retrieved (where R is the number of relevant objects)

## 4.3 Average precision (AP)

The measure [13] is a single-value that evaluates the performance over all relevant objects. It is not an average of the precision at standard recall levels, rather, it is the average of precision scores at each relevant object retrieved for example, consider a query that has five relevant objects which are retrieved at ranks 1,2,4,7,10. The actual precision obtained when each relevant object is retrieved is 1, 1, 0.75, 0.57, 0.50, respectively; the mean of them is 0.76.

## 4.4 Mean Average precision (MAP)

Find the average precision for each query and compute the mean of average precision [13] over all queries; it gives an overall evaluation of a retrieval algorithm.

## 4.5 E-Measures

The idea is to combine precision and recall into a single number to evaluation the whole system performance [12]. First we introduce the F-measure, which is the weighted harmonic mean of precision and recall. F-measure is defined as

$$F_\alpha = \frac{(1+\alpha) \times precision \times recall}{\alpha \times precision + recall} \text{, where } \alpha \text{ is the weight.} \qquad (2)$$

Let $\alpha$ be 1, the weight of precision and recall is same, and we have

$$F = 2 \times \frac{precision \times recall}{precisionl + recal} \qquad (3)$$

Then, go over all points on the precision-recall curve of each model and compute the F-measure, we get the overall evaluation of F for an algorithm.

The E-Measure is defined as E = 1- F,

$$E = 1 - \frac{2}{\frac{1}{P} + \frac{1}{R}} \quad (4)$$

Note that the maximum value is 1.0, and higher values indicate better results. The fact is that a user of a search engine is more interested in the first page of query results than in later pages. So, here we consider only the first 32 retrieved objects for every query and calculate the E-Measure over those results.

## 4.6 Discount Cumulative Gain (DCG)

Based on the idea that the greater the ranked position of a relevant object the less valuable it is for the user, because the less likely it is that the user will examine the object due to time, effort, and cumulated information from objects already seen.

In this evaluation, the relevance level of each object is used as a gained value measures for its ranked position *m,* the result and the gain is summed progressively from position 1 to n. Thus the ranked object lists (of some determined length) are turned to gained value lists by replacing object IDs with their relevance values. The binary relevance values 0, 1 are used (1 denoting relevant, 0 irrelevant) in our benchmark evaluation. Replace the object ID with the relevance values, we have for example:

G'=< 1, 1, 1, 0, 0, 1, 1,0,1,0 . . . . >

The cumulated gain at ranked position *i* is computed by summing from position 1 to *i* when *i* ranges from 1 to the length of the ranking list. Formally, let us denote position i in the gain vector G by G[i]. The cumulated gain vector CG is defined recursively as the vector CG where:

$$\mathrm{G[i]} = \begin{cases} \mathrm{G[1]} & \text{if } i = 1 \\ \mathrm{CG[i]} = \mathrm{CG[i-1]} + \mathrm{G[i]} & \text{otherwise} \end{cases} \quad (5)$$

The comparison of matching algorithms is then equal to compare the cumulated gain, the greater the rank, the smaller share of the object value is added to the cumulated gain. A discounting function is needed which progressively reduces the object weight as its rank increases but not too steeply:

$$\mathrm{DCG[i]} = \begin{cases} \mathrm{G[1]} & \text{if } i = 1 \\ \mathrm{DCG[i-1]} + \mathrm{G[i]} / \log_2 i & \text{otherwise} \end{cases} \quad (6)$$

The actual CG and DCG vectors by a particular matching algorithm may also be compared to the theoretically best possible. And this is called normalized CG, normalized DCG. The latter vectors are constructed as follows. Let there be 5 relevant objects, and 5 irrelevant objects in each class, then, at the relevance levels 0 and 1. Then the ideal Gain vector is obtain by filling the first vector positions with 1, and the remaining positions by the values 0. Then compute CG and DCG as well as the average CG and DCG vectors and curves as above. Note that the curves will turn horizontal when no more relevant objects (of any level) can be found. The vertical distance between an actual DCG/CG curve and the theoretically best possible curve shows the effort wasted on less-than-perfect objects due to a particular matching algorithm.

## 4.7 Nearest Neighbor (NN), First-tier (Tier1) and Second-tier (Tier2)

These evaluation measures [1] share the similar idea, that is, to check the ratio of models in the query's class that also appear within the top K matches, where K can be 1, the size of the query's class, or the double size of the query's class. Specifically, for a class with |C| members, K= 1 for Nearest Neighbor, K = |C| − 1 for the first tier, and K = 2 *(|C| − 1) for the second tier. In the NIST shape Benchmark database [6], C is always 20. The final score is an average over all the objects in database.

## 4.8 Computational Cost

For a number of vision based applications, such as Autonomous Robots, the speed of identification by different algorithms is very important. Computational cost is then related to the time it takes to extract the 3D shape descriptor for an object and perform one query search on the database, and the storage size (byte) of the shape descriptor.

# 5. THE NIST SHAPE BENCHMARK

In this section, we discuss the generic shape benchmark [6] constructed by our group. It contains 800 complete 3D models, which are categorized into 40 classes. The classes are defined with respect to their semantic categories. In each class there are 20 models. The NIST Shape Benchmark provides a new perspective in evaluating shape retrieval algorithms. It has several virtues: high reliability (in terms of error rate) to evaluate 3D shape retrieval algorithms, sufficient number of good quality models as the basis of the shape benchmark, equal size of classes to minimize the bias of evaluation.

## 5.1 Results

We present results of the ten algorithms that we tested on the generic benchmark. Table 1 compares different performance measures described in the previous section for different algorithms. Figure 1 Shows the overall Precision-recall curve for different algorithms on the new benchmark. In order to examine how different shape descriptors work on the database, we implement several kinds of algorithms to compare on the new benchmark. Moreover, comparison experiments are conducted on both the entire benchmark and a specific class of the benchmark. Several retrieval algorithms are evaluated from several aspects on this new benchmark by various measurements, and the reliability of the new shape benchmark.

Table 1: Retrieval performance of different algorithms on the NIST Shape Benchmark.

| | NN | Tier1 | Tier2 | E-Measure | DCG | MAP |
|---|---|---|---|---|---|---|
| LFD | 84.50% | 42.58% | 54.03% | 35.05% | 75.40% | 49.79% |
| Hybrid | 81.13% | 45.46% | 57.72% | 36.65% | 76.09% | 48.54% |
| EDT | 77.50% | 39.78% | 52.09% | 33.03% | 72.37% | 42.20% |
| DepthBuffer | 75.88% | 37.37% | 47.96% | 31.15% | 70.08% | 38.46% |
| SIL | 71.00% | 35.54% | 47.98% | 30.06% | 68.57% | 37.16% |
| MRSPRH | 70.00% | 35.15% | 45.99% | 29.98% | 68.77% | 36.38% |
| RSH | 69.13% | 32.83% | 44.11% | 28.18% | 66.43% | 37.16% |
| AAD | 65.00% | 30.61% | 41.69% | 26.67% | 64.58% | 31.59% |
| PS | 50.00% | 21.79% | 29.40% | 20.51% | 56.73% | 22.38% |
| D2 | 49.38% | 22.33% | 32.08% | 20.56% | 56.58% | 22.53% |

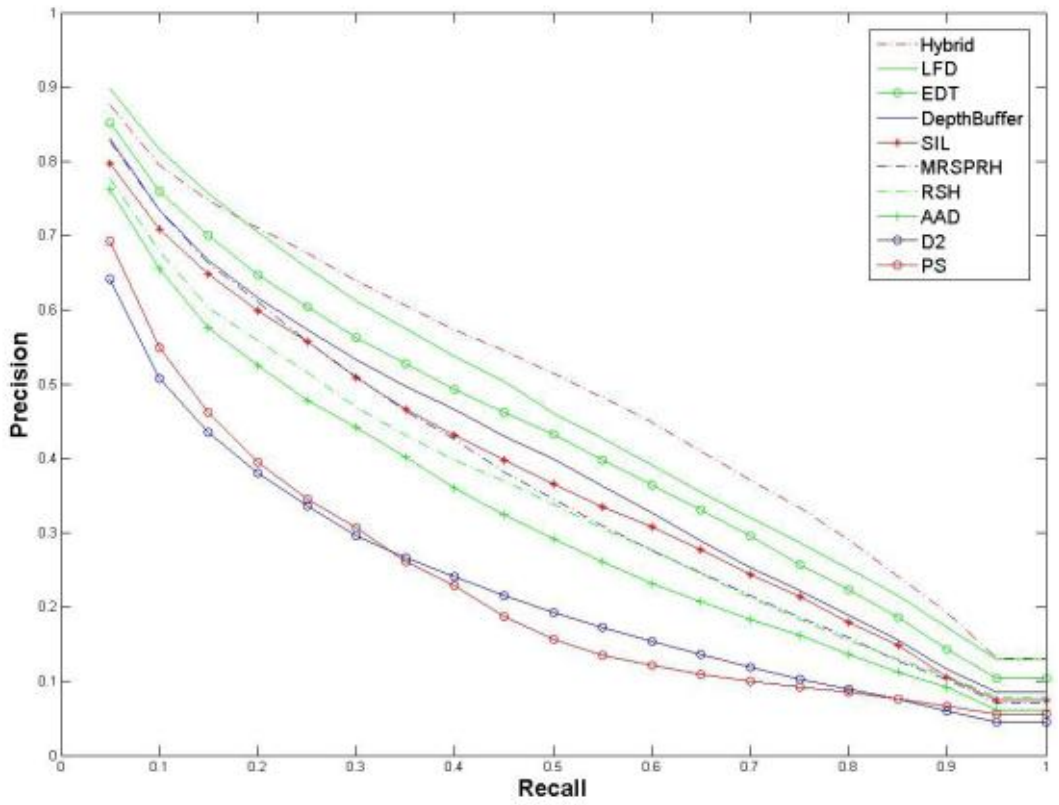

Figure 1: The overall Precision-recall curve for different algorithms on the NIST Shape Benchmark.

## 5.2 Reliability of a Benchmark

The reliability of a new proposed benchmark by testing the effect of class set size on retrieval error is an important issue. Voorhees and Buckley [17] proposed a method to estimate the reliability of retrieval experiments by computing the probability of making wrong decisions between two retrieval systems over two retrieval experiments. They also showed how the topic set sizes affect the reliability of retrieval experiments. We also conducted experiments to test the reliability of retrieval of the new generic 3D shape benchmark [6].

# 6. SHAPE RETRIEVAL CONTEST

In 2010 we organized three tracks in the 3D Shape Retrieval contest. The three tracks were the Generic 3D Warehouse Track [7], the Range scans Track [8], and the Non-rigid shapes Track [9]. These tracks were organized under the SHREC'10-3D Shape Retrieval Contest 2010 (www.aimatshape.net/event/SHREC), and in the context of the EuroGraphics 2010 Workshop On 3D Object Retrieval, 2010. SHREC’10 was the fifth edition of the contest. In the following subsections we will summarize the tracks that we organized.

## 6.1 Generic 3D Warehouse Contest

The aim of this track was to evaluate the performance of various 3D shape retrieval algorithms on a large Generic benchmark based on the Google 3D Warehouse. Three groups participated in the track and they submitted 7 set of results based on different methods and parameters. We also ran two standard algorithms on the dataset. The performance evaluation of this track is based on six different metrics described earlier. All the 3D models in the Generic 3D Warehouse track were acquired by a web crawler from the Google 3D Warehouse [19] which is an online collection of 3D models. The database consists of 3168 3D objects categorized into 43 categories. The number of objects in each category varies between 11 and 177. Figure 2 shows example of each category.

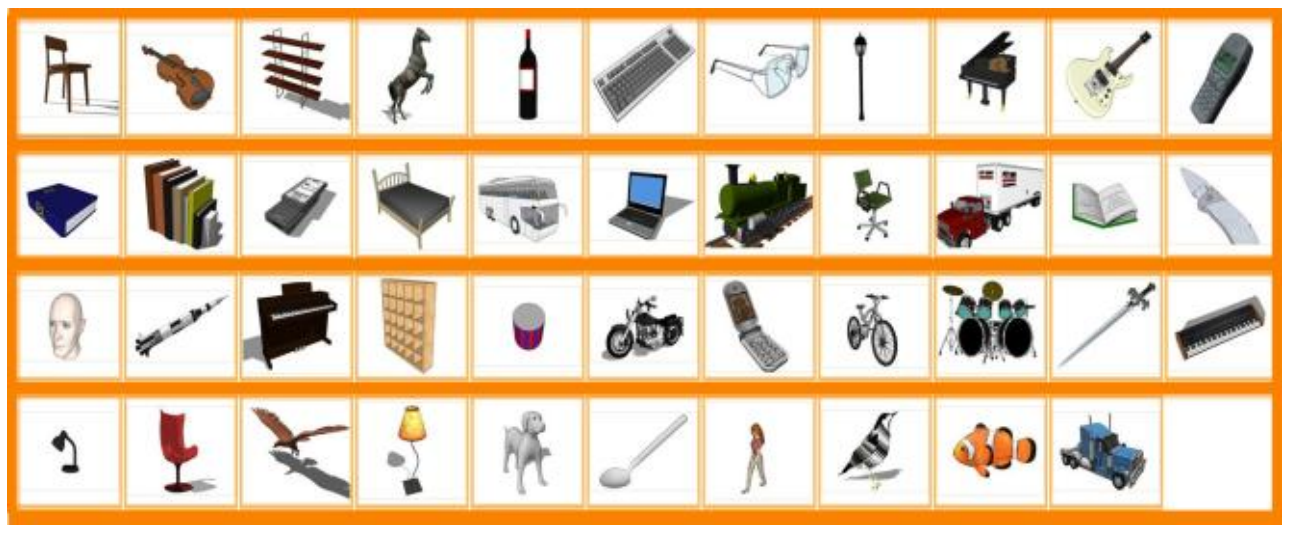

Figure 2: One example image from each class of the Generic 3D Warehouse Benchmark is shown.

### *6.1.1 Results*

In this section, we present the performance evaluation results of the Generic 3D Warehouse track. Table 2 shows the retrieval statistics yielded by the methods of the participants and five previous methods. Figure 3 gives the precision-recall curves of all the methods. Observing these figures, we can state that Lian’s VLGD+MMR method yielded highest results in terms of all the measures but Nearest Neighbor. While, in terms of Nearest Neighbor, Ohbuchi’s MR-BF-DSIFT-E method performed best.

Table 2: The retrieval statistics for all the methods and runs.

| PARTICIPANT | METHOD | NN | FT | ST | E | DCG |
|---|---|---|---|---|---|---|
| Ohbuchi | BF-DSIFT-E | 0.884 | 0.531 | 0.668 | 0.360 | 0.841 |
| | MR-BF-DSIFT-E | 0.897 | 0.606 | 0.733 | 0.389 | 0.869 |
| Dutagaci | View-based PCA - 18 view | 0.825 | 0.433 | 0.557 | 0.314 | 0.789 |
| Lian | GSMD | 0.875 | 0.491 | 0.624 | 0.344 | 0.824 |
| | CM-BOF | 0.862 | 0.534 | 0.662 | 0.358 | 0.836 |
| | VLGD | 0.889 | 0.565 | 0.696 | 0.377 | 0.855 |
| | VLGD+MMR | 0.889 | 0.647 | 0.791 | 0.390 | 0.880 |
| | | | | | | |
| AUTHOR | PREVIOUS METHODS | NN | FT | ST | E | DCG |
| Vranic | DSR472 with L1 | 0.871 | 0.498 | 0.639 | 0.356 | 0.831 |
| | DBD438 with L1 | 0.809 | 0.407 | 0.532 | 0.306 | 0.770 |
| | SIL300 with L1 | 0.807 | 0.412 | 0.548 | 0.300 | 0.780 |
| | RSH136 with L1 | 0.783 | 0.385 | 0.508 | 0.275 | 0.758 |
| Chen | LFD | 0.864 | 0.48 | 0.613 | 0.336 | 0.816 |

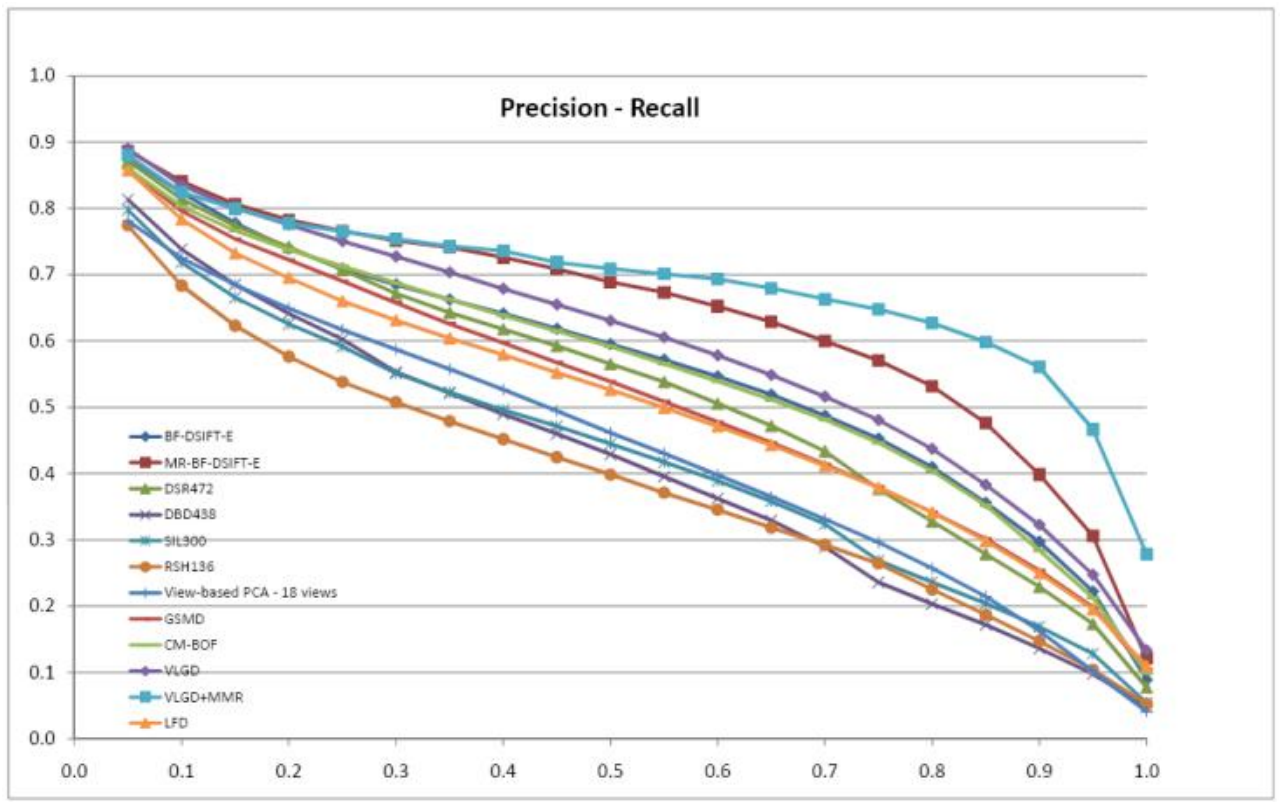

Figure 3: Precision-recall curves of the best runs of each participant.

## 6.2 Range Scan Retrieval Contest

In this contest, the aim was at comparing algorithms that match a range scan to complete 3D models in a target database. The queries are range scans of real objects, and the objective is to retrieve complete 3D models that are of the same class. The query set is composed of 120 range images, which are acquired by capturing 3 range scans of 40 real objects from arbitrary view directions, as shown in Figure 4. The target database is the generic shape benchmark constructed by our group [6]. It contains 800 complete 3D models, which are categorized into 40 classes

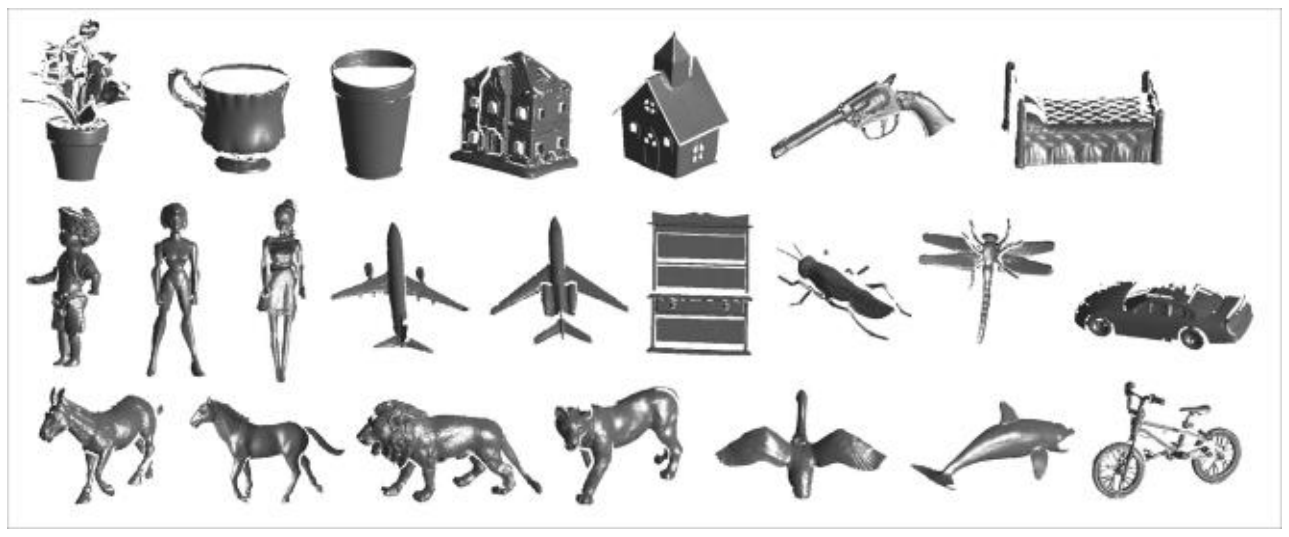

Figure 4: Examples from the query set.

### 6.2.1 Results

Two participants of the SHREC'10 track Range Scan Retrieval submitted five sets of rank lists each. The results for the ten submissions are summarized in the precision-recall curves in Figure 5. Figure 6 shows the models retrieved by one of the methods in response to a range scan of a toy bike.

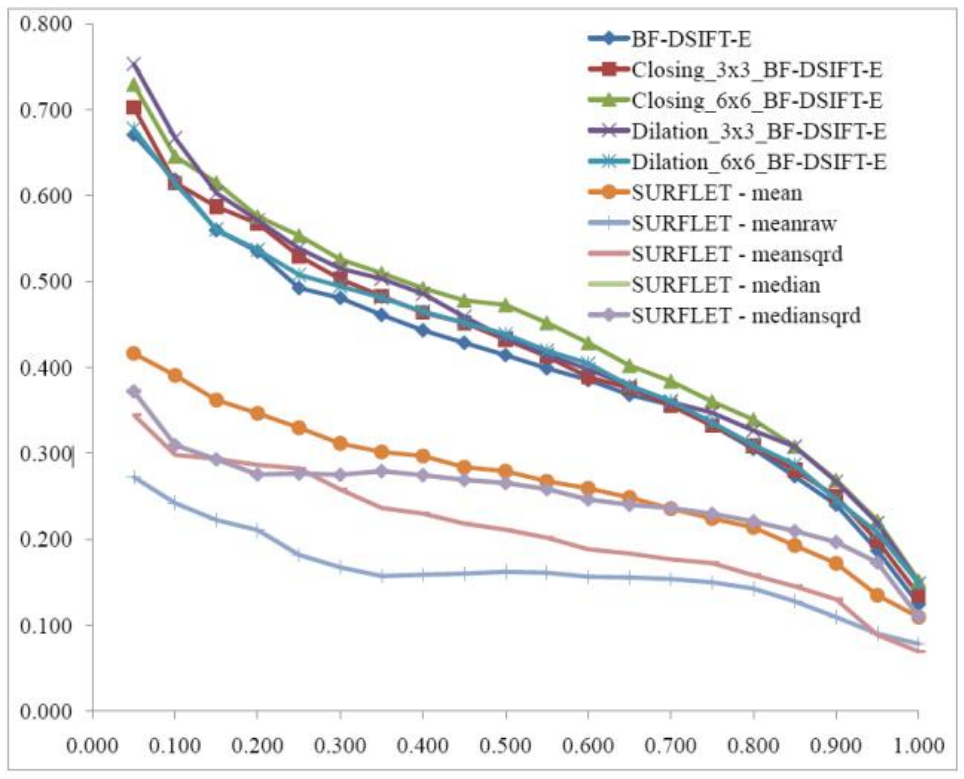

Figure 5: Precision-recall curves.

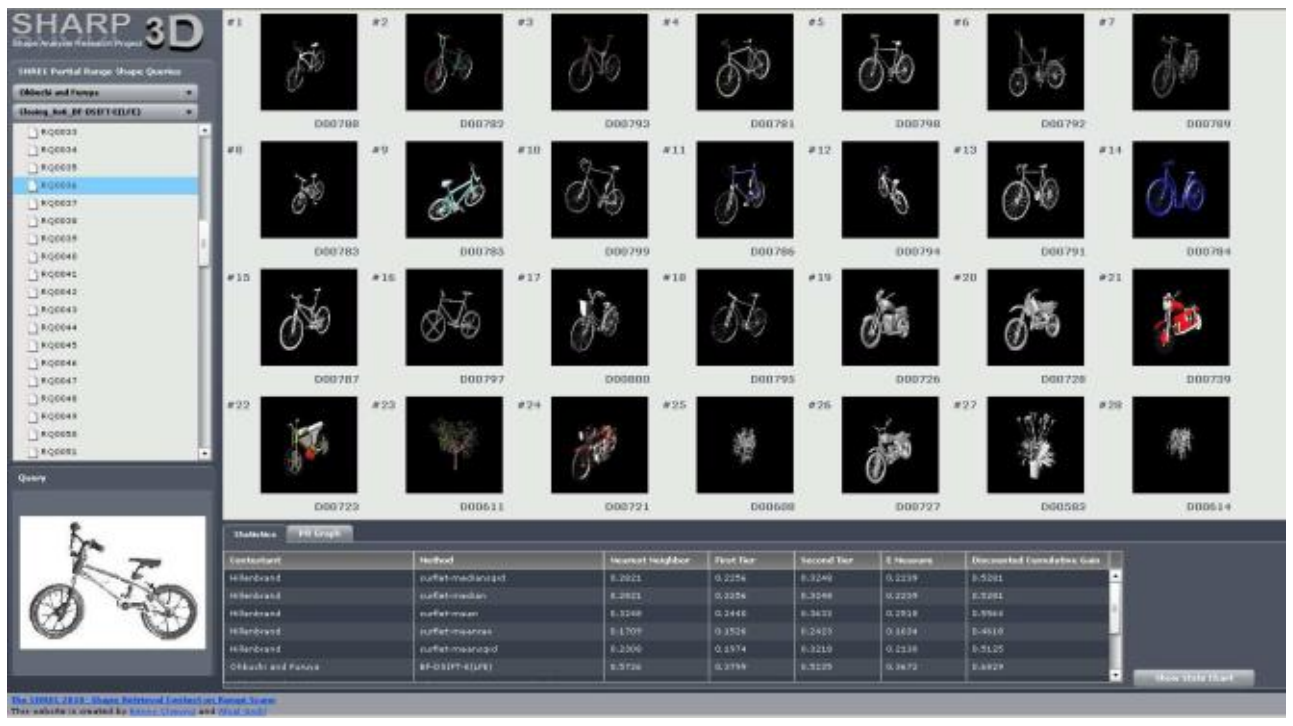

Figure 6: A sample shot from the web-based interface. The query is the range scan of a toy bike.

## 6.3 Non-rigid 3D Shape Retrieval Contest

The aim of this Contest was to evaluate and compare the effectiveness of different methods run on a non-rigid 3D shape benchmark consisting of 200 watertight triangular meshes. Three groups participated. The database used in this track consists of 200 watertight 3D triangular meshes, which are selected from the McGill Articulated [3] Shape Benchmark database.

### 6.3.1 Results

We present the results of the three groups that submitted six results. Figure 7 displays the Precision-recall curves to show retrieval performance of all six methods evaluated on the whole database. We also show the results using a web interface which displays the retrieved models for all objects and methods, to analyze the results as shown in Figure 8.

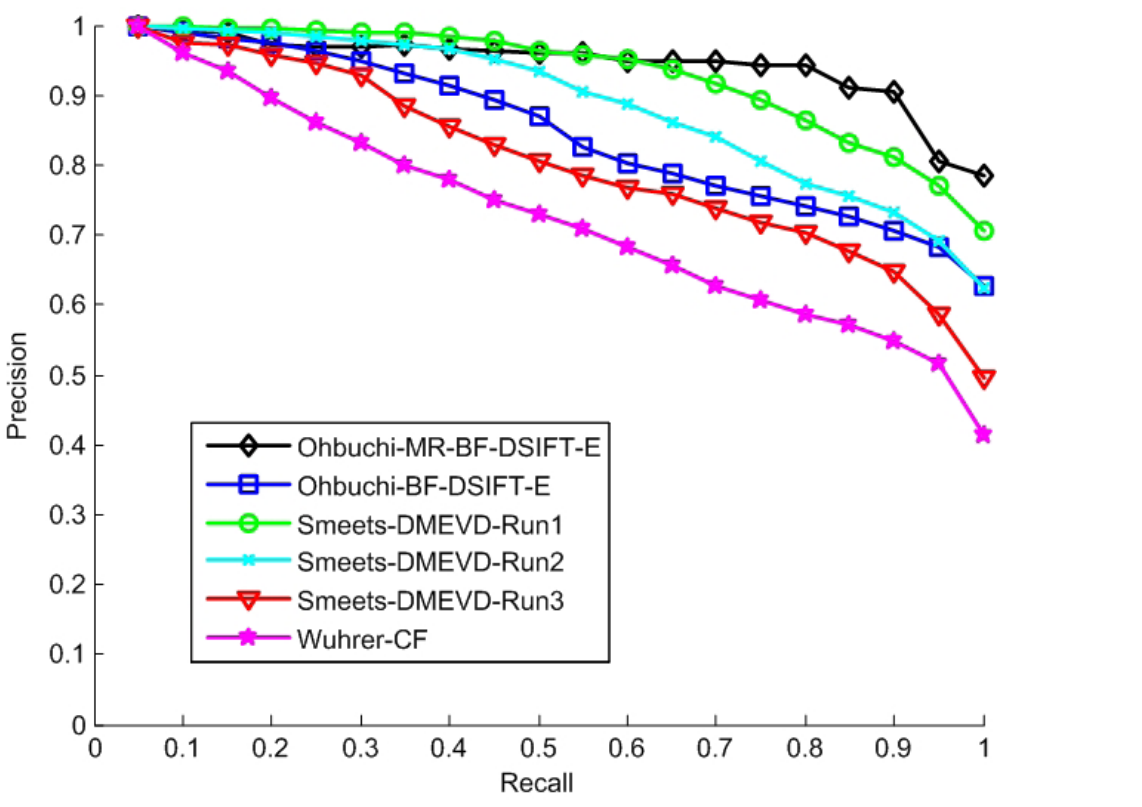

Figure 7: Precision-recall curves of all runs evaluated for the whole database.

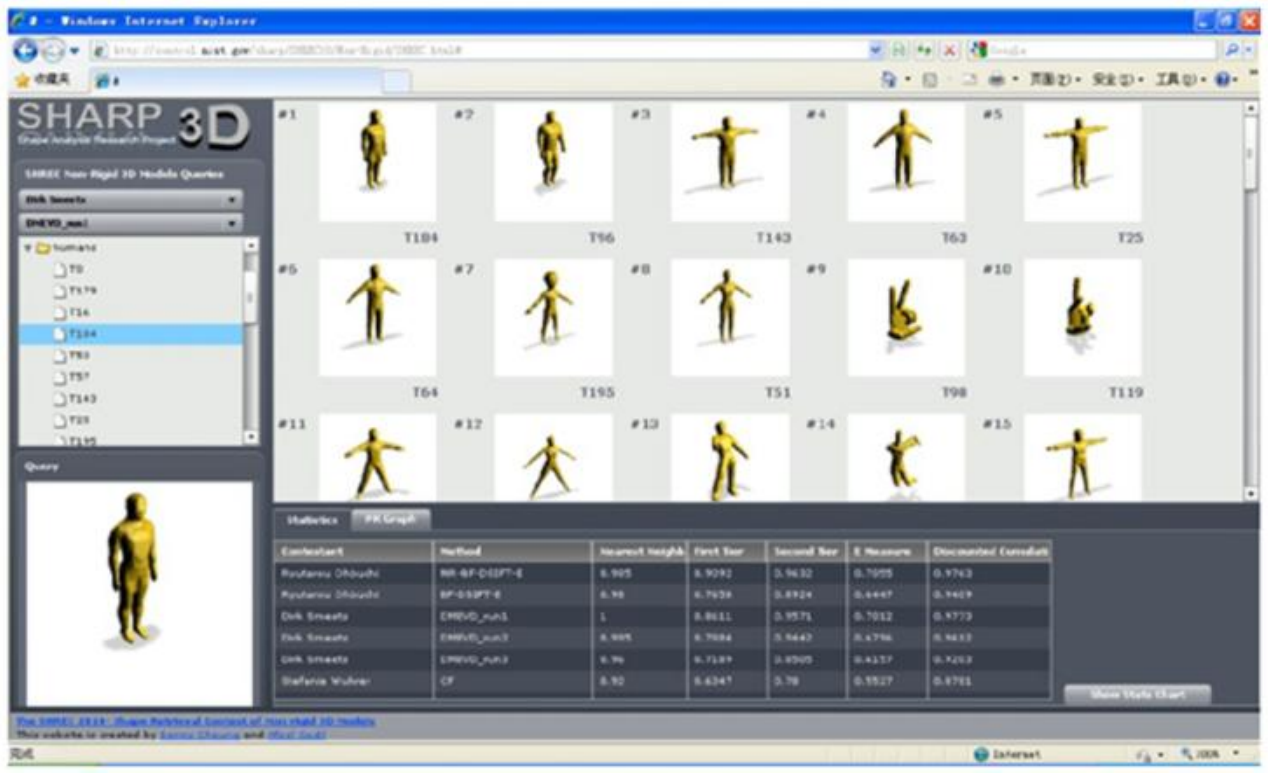

Figure 8. Retrieval example of one of the method using the web interface of the SHREC non-rigid track.

## 7. CONCLUSION

In this paper, we discussed some of the current 3D shape retrieval benchmarking efforts by our group and others and described the various steps involved in developing a benchmark. Then we reviewed the performance evaluation measures that are developed and used by researchers in the 3D shape retrieval community. We also gave an overview of the 3D shape retrieval contests (SHREC) run under the EuroGraphics Workshop on 3D Object Retrieval. Finally, we showed some of the results based on the NIST Shape benchmark and the different shape retrieval contest tracks we organized for SHREC 2010.

## 8. ACKNOWLEDGMENTS

This work has been supported by the SIMA and the IDUS program.